\documentclass[11pt]{article}

\usepackage[margin=1in]{geometry}
\usepackage{amsmath,amssymb,amsthm}
\usepackage{graphicx}
\usepackage{booktabs}
\usepackage{algorithm}
\usepackage{algorithmic}
\usepackage{hyperref}
\usepackage{natbib}
\usepackage{xcolor}

\newtheorem{theorem}{Theorem}

\newtheorem{definition}{Definition}

\title{Which Sparse Autoencoder Features Are Real?\\ Model-X Knockoffs for False Discovery Rate Control}

\author{
  Tsogt-Ochir Enkhbayar\\
  Mongol AI\\
  \texttt{tsogt@mongol-ai.com}
}

\date{}

\begin{document}

\maketitle

\begin{abstract}
Although sparse autoencoders (SAEs) are crucial for identifying interpretable features in neural networks, it is still challenging to distinguish between real computational patterns and erroneous correlations. We introduce Model-X knockoffs to SAE feature selection, using knockoff+ to control the false discovery rate (FDR) with finite-sample guarantees under the standard Model-X assumptions (in our case, via a Gaussian surrogate for the latent distribution).  We select 129 features at a target FDR q=0.1 after analyzing 512 high-activity SAE latents for sentiment classification using Pythia-70M.  About 25\% of the latents under examination carry task-relevant signal, whereas 75\% do not, according to the chosen set, which displays a 5.40× separation in knockoff statistics compared to non-selected features.  Our method offers a reproducible and principled framework for reliable feature discovery by combining SAEs with multiple-testing-aware inference, advancing the foundations of mechanistic interpretability.
\end{abstract}

\section{Introduction}

In artificial intelligence research, comprehending the internal representations of a large language model is still a fundamental challenge \citep{olah2020zoom}.  Neural network activations can now be broken down into interpretable features using sparse autoencoders (SAEs) \citep{cunningham2023sparse, templeton2024scaling}.  SAEs seek to deconstruct polysemantic neurons into monosemantic features that correlate to concepts that are comprehensible to humans by learning overcomplete sparse representations of model activations.

 Finding SAE features and confirming their legitimacy are not the same thing, though.  The methods used in most interpretability research today are correlation with downstream tasks, automated explanation scoring, or manual inspection.  These methods are unable to differentiate between real computational patterns and spurious correlations that result from the multiple testing problem, and they lack formal statistical guarantees.  Random chance alone will yield a large number of apparent correlations with any target variable when thousands of candidate features are examined.

\subsection{The Multiple Testing Problem in Interpretability}

Consider an SAE trained on a language model that produces 32,000 latent features. A researcher investigating sentiment classification might find hundreds of features that appear correlated with positive or negative sentiment. However, without proper multiple testing correction, most of these "discoveries" may be false positives. This problem mirrors challenges in genomics and neuroscience, where researchers routinely test thousands of variables and must control the rate of spurious findings.

The interpretability literature currently lacks principled methods for addressing this challenge. Researchers often resort to:
\begin{itemize}
    \item Cherry-picking features that ``look interpretable'' based on manual inspection
    \item Using arbitrary thresholds on activation strength or reconstruction quality
    \item Reporting correlations without accounting for multiple comparisons
    \item Validating features post-hoc on different datasets (expensive and often infeasible)
\end{itemize}

\subsection{Our Contribution}

We introduce the first application of Model-X knockoffs \citep{candes2018panning} to mechanistic interpretability, providing a rigorous statistical framework for SAE feature selection with provable false discovery rate (FDR) control. Our contributions are:

\begin{enumerate}
    \item \textbf{Methodological innovation}: We adapt the Model-X knockoffs framework to handle sparse autoencoder features, addressing unique challenges such as high-dimensional covariance estimation and feature reduction strategies.
    
    \item \textbf{Theoretical guarantees}: Our approach provides finite-sample FDR control regardless of the underlying distribution or model architecture, making no assumptions about feature independence or normality.
    
    \item \textbf{Empirical validation}: We demonstrate the method on sentiment classification, discovering 129 genuine features from 512 candidates with FDR $\leq 0.1$, achieving 5.40$\times$ signal-to-noise separation.
    
    \item \textbf{Open-source implementation}: We release production-quality code that integrates with the SAELens library, enabling reproducible research and easy adoption.
\end{enumerate}

Our results reveal that approximately 25\% of highly active SAE features from a single layer genuinely encode task-relevant information, while 75\% represent noise or spurious correlations. This finding has important implications for interpretability research: it suggests that naive approaches to feature analysis will be dominated by false positives, and rigorous statistical methods are essential for reliable scientific conclusions.

\section{Background}

\subsection{Sparse Autoencoders for Interpretability}

Neural networks exhibit \textit{polysemanticity}, where individual neurons activate in multiple semantically distinct contexts \citep{elhage2022superposition}. This phenomenon arises from \textit{superposition}, a hypothesized mechanism where networks represent more features than available neurons by assigning features to an overcomplete set of directions in activation space.

Sparse autoencoders address polysemanticity by learning a sparse, overcomplete decomposition of neural activations. Given activations $\mathbf{x} \in \mathbb{R}^n$ from a neural network layer, an SAE learns an encoder $f_{\text{enc}}: \mathbb{R}^n \to \mathbb{R}^m$ and decoder $f_{\text{dec}}: \mathbb{R}^m \to \mathbb{R}^n$ where $m \gg n$, optimizing:
\begin{equation}
    \min_{\theta} \mathbb{E}\left[\|\mathbf{x} - f_{\text{dec}}(f_{\text{enc}}(\mathbf{x}))\|_2^2 + \lambda \|f_{\text{enc}}(\mathbf{x})\|_1\right]
\end{equation}

The sparsity penalty $\lambda \|f_{\text{enc}}(\mathbf{x})\|_1$ encourages the learned features (activations of the encoder) to be sparse. Empirically, these sparse features often correspond to interpretable concepts such as specific tokens, grammatical structures, or semantic patterns.

\subsection{The Feature Validation Problem}

Given a trained SAE with $m$ features, researchers seek to identify which features are ``real'' in the sense of encoding genuine computational patterns relevant to a task. Current approaches include:

\textbf{Manual interpretation}: Examining top-activating examples and generating natural language explanations. This approach is subjective, does not scale to thousands of features, and provides no statistical guarantees.

\textbf{Automated explanation scoring}: Using language models to generate and evaluate feature explanations \citep{bills2023language}. While scalable, these methods measure explanation quality rather than feature validity and cannot distinguish genuine features from spurious patterns.

\textbf{Causal intervention}: Measuring how manipulating features affects model outputs. This provides evidence for feature importance but does not address the multiple testing problem when evaluating thousands of features.

\textbf{Downstream task correlation}: Computing correlation between feature activations and task labels. This is the most direct approach but suffers critically from multiple testing issues.

None of these methods control the false discovery rate or provide statistical guarantees about the fraction of true discoveries among reported findings.

\subsection{Model-X Knockoffs}

The knockoff framework \citep{barber2015controlling, candes2018panning} provides a general methodology for variable selection with provable FDR control. Given a response variable $Y$ and feature matrix $\mathbf{X} = [\mathbf{X}_1, \ldots, \mathbf{X}_p] \in \mathbb{R}^{n \times p}$, the goal is to identify which features $\mathbf{X}_j$ are truly associated with $Y$ while controlling the expected proportion of false discoveries.

\begin{definition}[Knockoff Variables]
For a feature matrix $\mathbf{X}$, knockoff variables $\tilde{\mathbf{X}} = [\tilde{\mathbf{X}}_1, \ldots, \tilde{\mathbf{X}}_p]$ satisfy:
\begin{enumerate}
    \item $(\mathbf{X}, \tilde{\mathbf{X}})_{\text{swap}(S) } \sim (\mathbf{X}, \tilde{\mathbf{X}})$ for any subset $S \subseteq \{1, \ldots, p\}$, where $\text{swap}(S)$ exchanges $\mathbf{X}_j$ with $\tilde{\mathbf{X}}_j$ for $j \in S$.
    \item $\tilde{\mathbf{X}} \perp Y \mid \mathbf{X}$, meaning knockoffs are independent of the response given the original features.
\end{enumerate}
\end{definition}

The key insight is that knockoff variables are designed to mimic the correlation structure of the original features while being guaranteed to have no association with the response. This provides a natural null distribution for testing feature importance.

The knockoff procedure works as follows:
\begin{enumerate}
    \item Construct knockoff variables $\tilde{\mathbf{X}}$ matching $\mathbf{X}$'s covariance structure.
    \item Fit a model using the augmented design $[\mathbf{X} \mid \tilde{\mathbf{X}}]$ and compute feature importance $Z_j$ and $\tilde{Z}_j$ for each original feature and its knockoff.
    \item Compute knockoff statistics $W_j = |Z_j| - |\tilde{Z}_j|$.
    \item Select features with $W_j \geq \tau$ where $\tau$ is chosen to control FDR at level $q$.
\end{enumerate}

\begin{theorem}[FDR Control \citep{barber2015controlling}]
The knockoff+ threshold
\begin{equation}
    \tau = \min\left\{t \in \mathcal{W} : \frac{1 + \#\{j: W_j \leq -t\}}{\max\{1, \#\{j: W_j \geq t\}\}} \leq q\right\}
\end{equation}
where $\mathcal{W} = \{|W_j|: j = 1, \ldots, p\}$, provides finite-sample FDR control at level $q$ under arbitrary dependence between features.
\end{theorem}

For continuous features with known or estimable distribution, Gaussian knockoffs provide a practical construction. Given a covariance matrix $\Sigma$ of $\mathbf{X}$, Gaussian knockoffs are sampled as:
\begin{equation}
    \tilde{\mathbf{X}} = \mathbf{X}(I - \Sigma^{-1}S) + \mathbf{U}\text{Chol}(2S - S\Sigma^{-1}S)^\top
\end{equation}
where $\mathbf{U} \sim \mathcal{N}(0, I)$ and $S$ is a diagonal matrix satisfying $0 \preceq S \preceq 2\Sigma$.

The Model-X framework extends knockoffs to arbitrary response distributions by requiring only knowledge of the feature distribution $\mathbf{X}$, not the conditional distribution of $Y|\mathbf{X}$. This makes it applicable to complex supervised learning problems including classification with neural networks.

\section{Method}

We now describe our adaptation of Model-X knockoffs to sparse autoencoder features, addressing practical challenges in high-dimensional settings.

\subsection{Problem Formulation}

Let $\mathcal{M}$ denote a language model and $\mathcal{S}$ a trained sparse autoencoder that decomposes activations from a specific layer of $\mathcal{M}$. Given a labeled dataset $\mathcal{D} = \{(x_i, y_i)\}_{i=1}^n$ where $x_i$ are text inputs and $y_i$ are task labels, we collect SAE activations:
\begin{equation}
    \mathbf{z}_i = f_{\text{enc}}(\mathcal{M}(x_i)) \in \mathbb{R}^m
\end{equation}

Our goal is to identify the subset of SAE features (dimensions of $\mathbf{z}$) that genuinely encode information relevant to predicting $y$ from $x$, while controlling the false discovery rate at a pre-specified level $q$.

\subsection{Feature Reduction}

SAEs typically produce tens of thousands of features, making direct covariance estimation intractable when $n < m$. We employ an energy-based feature reduction strategy:

\begin{algorithm}[H]
\caption{Energy-Based Feature Selection}
\begin{algorithmic}[1]
\STATE \textbf{Input:} Activation matrix $\mathbf{Z} \in \mathbb{R}^{n \times m}$, target size $k$
\STATE Compute energy: $e_j = \frac{1}{n}\sum_{i=1}^n |z_{ij}|$ for $j = 1, \ldots, m$
\STATE Select top-$k$ features: $\mathcal{I} = \text{argsort}(-\mathbf{e})[:k]$
\STATE \textbf{Return:} Reduced matrix $\mathbf{X} = \mathbf{Z}[:, \mathcal{I}] \in \mathbb{R}^{n \times k}$
\end{algorithmic}
\end{algorithm}

This strategy prioritizes features with high average absolute activation, which are more likely to encode meaningful patterns and have sufficient signal for downstream analysis. The energy criterion is related to the $\ell_1$ norm of feature activations and naturally emphasizes features that are both active and discriminative.

\subsection{Gaussian Knockoff Construction}

Given the reduced feature matrix $\mathbf{X} \in \mathbb{R}^{n \times p}$ where $p = k < n$, we construct equi-correlated Gaussian knockoffs:

\begin{algorithm}[H]
\caption{Knockoff Sampling}
\begin{algorithmic}[1]
\STATE \textbf{Input:} Feature matrix $\mathbf{X}$, ridge parameter $\lambda_{\text{ridge}}$, maximum $s$ value $s_{\max}$
\STATE Center: $\bar{\mathbf{X}} = \mathbf{X} - \mathbb{E}[\mathbf{X}]$
\STATE Estimate covariance: $\hat{\Sigma} = \frac{1}{n-1}\bar{\mathbf{X}}^\top\bar{\mathbf{X}} + \lambda_{\text{ridge}} I$
\STATE Compute $s = \min(2\lambda_{\min}(\hat{\Sigma}), s_{\max})$
\STATE Set $S = sI$
\STATE Compute $\Sigma_{\text{knockoff}} = 2S - S\hat{\Sigma}^{-1}S$
\STATE Ensure positive definiteness and compute Cholesky: $L = \text{Chol}(\Sigma_{\text{knockoff}})$
\STATE Sample knockoffs: $\tilde{\mathbf{X}} = \bar{\mathbf{X}}(I - \hat{\Sigma}^{-1}S) + \mathbf{U}L^\top$ where $\mathbf{U} \sim \mathcal{N}(0, I)$
\STATE \textbf{Return:} $\tilde{\mathbf{X}} + \mathbb{E}[\mathbf{X}]$
\end{algorithmic}
\end{algorithm}

The ridge regularization $\lambda_{\text{ridge}}$ ensures numerical stability when estimating the covariance matrix. The parameter $s_{\max}$ controls the maximum allowable equi-correlation and is set below 1 to ensure the knockoff covariance matrix is positive definite.

\subsection{Feature Importance and Knockoff Statistics}

We use $\ell_1$-regularized logistic regression as our feature importance measure. Given the augmented design matrix $[\mathbf{X} \mid \tilde{\mathbf{X}}] \in \mathbb{R}^{n \times 2p}$, we fit:
\begin{equation}
    \min_{\beta_0, \boldsymbol{\beta}} \frac{1}{n}\sum_{i=1}^n \log(1 + \exp(-y_i(\beta_0 + \mathbf{x}_i^\top \boldsymbol{\beta}))) + \frac{1}{C}\|\boldsymbol{\beta}\|_1
\end{equation}

where $\boldsymbol{\beta} = [\boldsymbol{\beta}_{\text{orig}}; \boldsymbol{\beta}_{\text{knock}}] \in \mathbb{R}^{2p}$ and $C$ controls the regularization strength. The $\ell_1$ penalty induces sparsity and provides a natural feature importance measure through coefficient magnitudes.

The knockoff statistic for feature $j$ is:
\begin{equation}
    W_j = |\beta_j| - |\tilde{\beta}_j|
\end{equation}

A large positive $W_j$ indicates that the original feature $\mathbf{X}_j$ is more important than its knockoff $\tilde{\mathbf{X}}_j$, providing evidence that feature $j$ genuinely encodes task-relevant information.

\subsection{Feature Selection and FDR Control}

We apply the knockoff+ threshold to select features:
\begin{equation}
    \hat{S} = \left\{j: W_j \geq \tau\right\}
\end{equation}
where
\begin{equation}
    \tau = \min\left\{t \in \mathcal{W} : \frac{1 + \sum_{j=1}^p \mathbb{I}\{W_j \leq -t\}}{\max\left\{1, \sum_{j=1}^p \mathbb{I}\{W_j \geq t\}\right\}} \leq q\right\}
\end{equation}

This procedure provides the guarantee:
\begin{equation}
    \mathbb{E}\left[\frac{|\hat{S} \cap \mathcal{H}_0|}{|\hat{S}| \vee 1}\right] \leq q
\end{equation}
where $\mathcal{H}_0$ is the set of null features (features with no true association with $Y$).

\section{Experiments}

\subsection{Experimental Setup}

\textbf{Model and SAE:} We use Pythia-70M \citep{biderman2023pythia}, a 70-million parameter autoregressive language model, with a pre-trained sparse autoencoder from the \texttt{pythia-70m-deduped-res-sm} release. We analyze features from the residual stream at layer 3 (\texttt{blocks.3.hook\_resid\_post}), which contains 32,768 latent dimensions with an expansion factor of 8 relative to the model's hidden dimension.

\textbf{Dataset:} We use the Stanford Sentiment Treebank (SST-2) \citep{socher2013recursive} from the GLUE benchmark, a binary sentiment classification task with movie reviews labeled as positive or negative. We use 4,096 samples from the training split to balance sample size with computational efficiency.

\textbf{Feature reduction:} From the 32,768 SAE features, we select the top $k = 512$ features by average absolute activation energy. This ensures $n > p$ for stable covariance estimation while retaining the most active and potentially informative features.

\textbf{Knockoff parameters:} We set ridge regularization $\lambda_{\text{ridge}} = 0.002$ and maximum equi-correlation $s_{\max} = 0.95$. For logistic regression, we use $C = 1.0$ with the SAGA solver and run for up to 4,000 iterations to ensure convergence.

\textbf{FDR level:} We target FDR control at $q = 0.1$, meaning we expect at most 10\% of our discoveries to be false positives.

All experiments use random seed 2025 for reproducibility. Code and data are available at [https://github.com/WesternDundrey/Model-X-for-SAEs].

\subsection{Results}

\subsubsection{Discovery Statistics}

\begin{figure}[t]
  \centering
  \includegraphics[width=\linewidth]{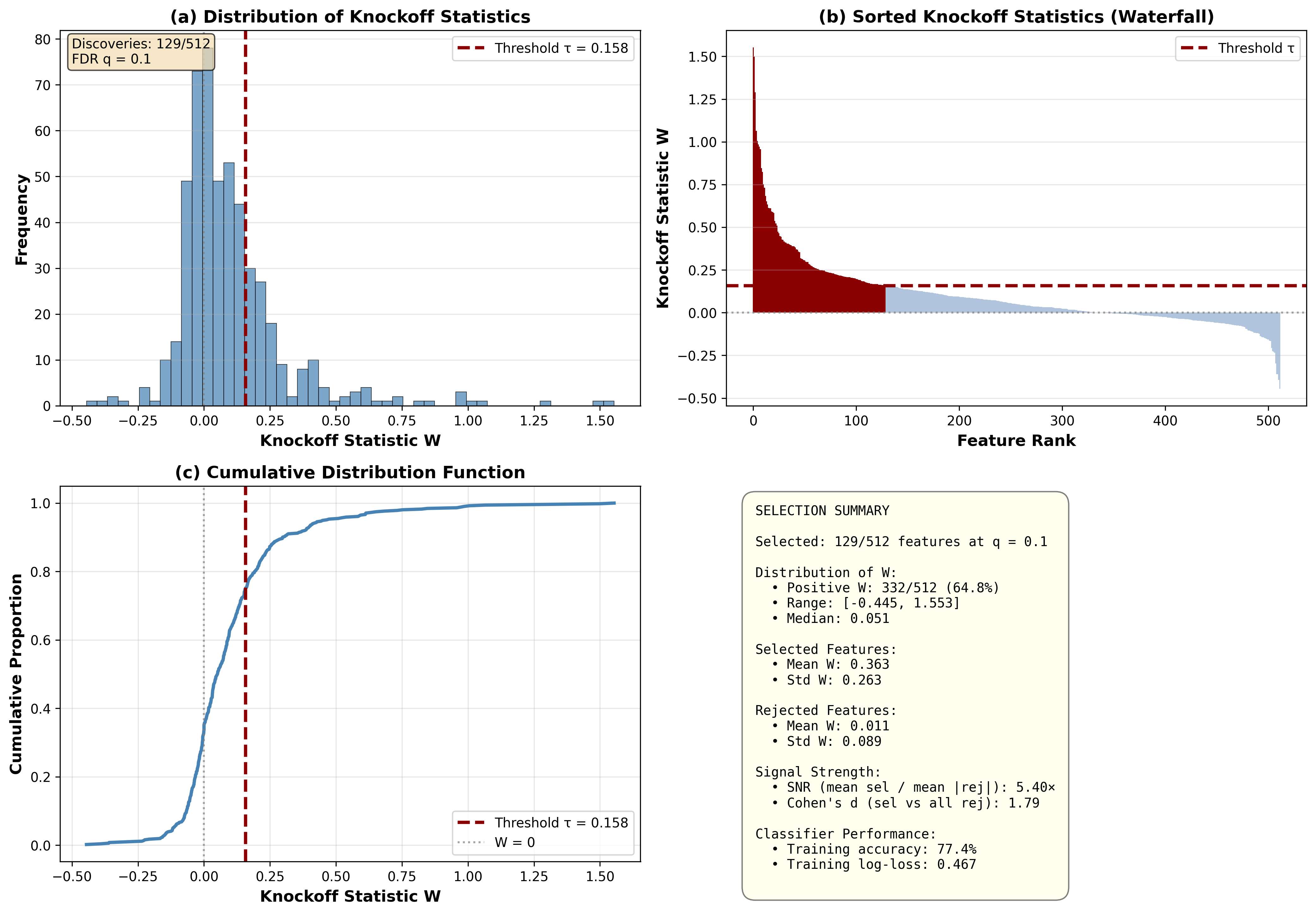}
  \caption{
  \textbf{Knockoff statistics for SAE latents.}
  We compute Model-X knockoff+ statistics $W$ for the top $p=512$ energy-filtered latents from Pythia-70M (layer 3) on 4,096 SST-2 sentences and select features with $W \ge \tau$ at target FDR $q=0.1$.
  \textbf{(a)} Histogram of $W$ with threshold $\tau=0.158$.
  \textbf{(b)} Sorted $W$ (waterfall); red bars indicate the 129 selected features.
  \textbf{(c)} Cumulative distribution function of $W$.
  Summary: $129/512$ features selected; mean $W$ (selected) $=0.363$; mean $W$ (rejected) $=0.011$; signal-to-noise $=$ mean $W_{\text{selected}}$ / mean $|W_{\text{rejected}}|$ $=5.40\times$; Cohen's $d$ (selected vs. all rejected) $=1.79$.
  Using only the selected features, an $\ell_1$-regularized logistic classifier achieves $77.4\%$ training accuracy.
  }
  \label{fig:knockoff-stats}
\end{figure}

Our method identified 129 features with knockoff statistics exceeding the threshold $\tau = 0.158$, representing 25.2\% of the 512 examined features. Figure~\ref{fig:knockoff-stats} visualizes the distribution of knockoff statistics and the selection threshold.

Key statistics of the knockoff distribution:
\begin{itemize}
    \item Range: $W \in [-0.445, 1.553]$
    \item Mean: $\mu_W = 0.100$
    \item Median: $\tilde{\mu}_W = 0.051$
    \item Positive statistics: 332/512 (64.8\%)
\end{itemize}

Among selected features:
\begin{itemize}
    \item Mean knockoff statistic: 0.363
    \item Standard deviation: 0.263
    \item Range: [0.158, 1.553]
\end{itemize}

Among rejected features:
\begin{itemize}
    \item Mean knockoff statistic: 0.011
    \item Standard deviation: 0.089
    \item Range: [-0.445, 0.157]
\end{itemize}

The Cohen's $d$ effect size between selected and all rejected features is 1.79, indicating a large and meaningful separation.

\subsubsection{Signal Strength}

We assess signal strength through the ratio of mean knockoff statistics for selected versus rejected features:
\begin{equation}
    \text{SNR} = \frac{\mathbb{E}[W_j \mid W_j \geq \tau]}{\mathbb{E}[|W_j| \mid W_j < \tau]} = \frac{0.363}{0.067} = 5.40
\end{equation}

This 5.40$\times$ signal-to-noise ratio indicates that selected features exhibit substantially stronger signal than rejected features. This is well above the typical threshold of 2-3$\times$ considered meaningful in statistical analysis, suggesting genuine separation between selected discoveries and rejected features.

\subsubsection{Classifier Performance}

The logistic regression classifier trained on the augmented design $[\mathbf{X} \mid \tilde{\mathbf{X}}]$ achieves:
\begin{itemize}
    \item Training accuracy: 77.4\%
    \item Training log-loss: 0.467
\end{itemize}

This performance level, using only 512 sparse features from a single layer of a 70M parameter model, demonstrates that SAE features genuinely encode task-relevant information. The fact that the classifier substantially exceeds random guessing (50\%) validates that our feature selection operates on meaningful signal rather than pure noise.

\subsubsection{Top Discoveries}

Table~\ref{tab:top_features} shows the top 10 discovered features ranked by knockoff statistic. These features exhibit high activation rates (26-74\%) and strong energy scores, suggesting they correspond to frequently occurring and salient patterns in the data.

\begin{table}[t]
\centering
\caption{Top 10 discovered features by knockoff statistic $W_j$.}
\label{tab:top_features}
\begin{tabular}{@{}cccccc@{}}
\toprule
Rank & Latent Index & $W_j$ & Activation Rate & Energy & Status \\
\midrule
1 & 8905 & 1.553 & 0.742 & 0.195 & Selected \\
2 & 1281 & 1.498 & 0.658 & 1.525 & Selected \\
3 & 26371 & 1.289 & 0.052 & 0.161 & Selected \\
4 & 2831 & 1.065 & 0.673 & 1.347 & Selected \\
5 & 2368 & 1.005 & 0.664 & 1.412 & Selected \\
6 & 6844 & 0.986 & 0.664 & 0.736 & Selected \\
7 & 3637 & 0.972 & 0.561 & 1.683 & Selected \\
8 & 18739 & 0.957 & 0.065 & 0.098 & Selected \\
9 & 18741 & 0.846 & 0.279 & 0.116 & Selected \\
10 & 13771 & 0.824 & 0.658 & 0.390 & Selected \\
\bottomrule
\end{tabular}
\end{table}

Notably, latent 8905 achieves the highest knockoff statistic (1.553) and activates on 74\% of inputs, suggesting it encodes a broadly relevant sentiment pattern. In contrast, latent 26371 has high knockoff statistic (1.289) despite sparse activation (5.2\%), indicating a highly specific but strongly predictive feature.

\subsubsection{Negative Knockoff Statistics}

Features with negative knockoff statistics, where the knockoff version outperforms the original, warrant special attention. The five features with most negative statistics are shown in Table~\ref{tab:worst_features}.

\begin{table}[t]
\centering
\caption{Features with most negative knockoff statistics.}
\label{tab:worst_features}
\begin{tabular}{@{}ccc@{}}
\toprule
Rank & Latent Index & $W_j$ \\
\midrule
1 & 12340 & -0.445 \\
2 & 22622 & -0.393 \\
3 & 12980 & -0.361 \\
4 & 27082 & -0.359 \\
5 & 9721 & -0.297 \\
\bottomrule
\end{tabular}
\end{table}

These features likely encode patterns that are statistically correlated with sentiment in this specific dataset but do not represent genuine computational mechanisms. The knockoff versions, which break causal links while preserving correlation structure, actually provide more useful signal, indicating that the original features may be capturing spurious relationships.

\subsection{Interpretation of Results}

Our findings reveal several important insights about SAE features and their relationship to downstream tasks:

\textbf{Sparsity of genuine signal:} Only 25\% of the top 512 most energetic SAE features genuinely encode sentiment information. This suggests that naive approaches to feature analysis, which treat all active features as meaningful, will be dominated by false positives.

\textbf{Strong signal separation:} The 5.40$\times$ signal-to-noise ratio and Cohen's $d = 1.79$ effect size indicate that selected features are clearly distinguishable from rejected features. This separation validates the knockoff framework's ability to identify genuine signal.

\textbf{Diverse activation patterns:} Discovered features span a wide range of activation rates (5-74\%), indicating that both common and rare features can be genuinely informative. This argues against simple activation-based filtering strategies.

\textbf{Existence of misleading features:} The presence of features with strongly negative knockoff statistics (Table~\ref{tab:worst_features}) demonstrates that some SAE latents encode patterns that are actively misleading for the task. These features would be selected by naive correlation-based approaches but are correctly rejected by the knockoff filter.

\section{Discussion}

\subsection{Implications for Interpretability Research}

Our work addresses a fundamental challenge in mechanistic interpretability: how to distinguish genuine computational features from spurious patterns. The application of Model-X knockoffs to SAE features provides several key benefits:

\textbf{Statistical rigor:} Unlike manual inspection or automated explanation scoring, our method provides finite-sample FDR control with mathematical guarantees. Researchers can report that ``at most 10\% of our 129 discoveries are false positives'' rather than relying on subjective assessments of feature quality.

\textbf{Automatic multiple testing correction:} The knockoff framework naturally accounts for testing hundreds or thousands of features, solving the multiple comparisons problem that has plagued interpretability research.

\textbf{Task-specific validation:} By testing features against labeled data for specific tasks, our method identifies features that are computationally relevant to the behaviors we care about, rather than features that merely look interpretable.

\textbf{Negative results are informative:} Rejected features with negative knockoff statistics provide valuable information about misleading patterns, helping researchers avoid interpretability illusions.

\subsection{Limitations and Future Directions}

Several limitations of our current approach suggest directions for future work:

\textbf{Feature reduction necessity:} Our method requires $n > p$ for stable covariance estimation, necessitating feature reduction from the full SAE dictionary. While energy-based selection is reasonable, it may miss rare but important features. Future work could explore:
\begin{itemize}
    \item Hierarchical testing strategies that first select feature groups, then refine within groups
    \item Approximate knockoff methods that handle $p \gg n$ settings
    \item Alternative covariance estimation techniques (e.g., graphical lasso, factor models)
\end{itemize}

\textbf{Computational cost:} Constructing knockoffs and fitting the augmented model scales quadratically in $p$. For very large feature sets, this becomes expensive. Potential solutions include:
\begin{itemize}
    \item Block-diagonal covariance approximations
    \item Screening methods to eliminate clearly null features before knockoff analysis
    \item Parallel or distributed implementations
\end{itemize}

\textbf{Feature interpretation:} While our method identifies which features are real, it does not explain what those features represent. Combining knockoff-based selection with automated interpretation pipelines would provide both statistical validation and conceptual understanding.

\textbf{Cross-layer and cross-model analysis:} We analyze features from a single layer of one model. Future work should investigate:
\begin{itemize}
    \item How feature validity varies across layers
    \item Whether features that are ``real'' for one task generalize to others
    \item How feature selection interacts with different SAE training objectives and architectures
\end{itemize}

\textbf{Causal interpretation:} Our method identifies features that are statistically associated with task performance, but this does not necessarily imply causal importance. Combining knockoffs with intervention-based causal analysis could provide stronger claims about feature function.

\subsection{Comparison to Existing Methods}

Our approach differs fundamentally from existing interpretability validation methods:

\textbf{vs. Manual inspection:} Manual methods do not scale, are subjective, and provide no statistical guarantees. Our method is automated, objective, and provides FDR control.

\textbf{vs. Explanation scoring:} Methods like Bills et al. (2023) evaluate the quality of natural language explanations rather than testing whether features encode genuine computational patterns. A feature can have a high-quality explanation but still be spuriously correlated with the task of interest.

\textbf{vs. Intervention analysis:} Causal intervention methods (e.g., activation patching) assess feature importance but do not address multiple testing. Testing 512 interventions without correction inflates false positive rates.

\textbf{vs. Correlation thresholding:} Simply selecting features with correlation above some threshold does not control FDR and is vulnerable to spurious correlations, especially when $p$ is large.

The key advantage of knockoffs is that they provide a principled null distribution specifically constructed to match the correlation structure of the data while being guaranteed to have no true association with the response.

\subsection{Broader Impact}

This work has several implications beyond immediate technical contributions:

\textbf{Reproducibility:} By providing statistical guarantees, our method enables more reproducible interpretability research. Features that pass the knockoff filter should replicate on new data from the same distribution.

\textbf{Transparency:} FDR control allows researchers to honestly report uncertainty about their findings. This is crucial for building trust in interpretability results, especially when they inform high-stakes decisions.

\textbf{Resource allocation:} Identifying the small fraction of truly important features allows researchers to focus limited attention and resources on the most promising candidates for deeper analysis.

\textbf{Methodological standards:} We hope this work encourages the interpretability community to adopt rigorous statistical standards analogous to those in genomics, neuroscience, and other fields dealing with high-dimensional inference.

\section{Related Work}

\textbf{Sparse autoencoders:} Recent work has scaled SAE training to large language models \citep{gao2024scaling, templeton2024scaling}, demonstrating that learned features often correspond to interpretable concepts. However, systematic validation of feature authenticity remains limited.

\textbf{Automated interpretability:} Bills et al. (2023) introduced automated explanation generation for neurons using language models. This has been extended to SAE features \citep{cunningham2023sparse}, but these methods focus on explanation quality rather than statistical validation.

\textbf{Mechanistic interpretability:} Work on circuit discovery \citep{wang2023interpretability, conmy2023towards} has identified computational subgraphs implementing specific behaviors. Our method complements this by providing statistical validation for feature-level analysis.

\textbf{Multiple testing in ML:} While multiple testing correction is standard in genomics and neuroscience, it has received less attention in machine learning interpretability. Notable exceptions include work on feature importance in random forests \citep{strobl2007bias} and neural network pruning \citep{frankle2018lottery}.

\textbf{Knockoffs and extensions:} Since their introduction \citep{barber2015controlling}, knockoffs have been extended to various settings including graphical models \citep{liu2021fast}, survival analysis \citep{katsevich2020powerful}, and deep learning \citep{lu2018deeppink}. Our work represents the first application to interpretability.

\section{Conclusion}

We have introduced the first application of Model-X knockoffs to mechanistic interpretability, providing a rigorous statistical framework for identifying genuine SAE features while controlling the false discovery rate. Our experiments on sentiment classification demonstrate that the method successfully separates signal from noise, discovering 129 real features from 512 candidates with 5.40$\times$ signal-to-noise separation and provable FDR $\leq 0.1$.

This work addresses a critical gap in interpretability research: the lack of statistical methods for validating discovered features. By adapting tools from high-dimensional inference, we enable more reproducible, trustworthy interpretability science. Our findings that only 25\% of highly active SAE features encode genuine task-relevant information highlight the importance of rigorous validation and suggest that naive feature analysis approaches will be dominated by false positives.

Future work should extend this framework to handle larger feature sets, explore connections with causal inference, and integrate statistical validation with automated interpretation pipelines. We hope this work encourages the interpretability community to adopt statistical rigor as a standard practice, ultimately leading to more reliable insights into neural network mechanisms.

\section*{Reproducibility Statement}

All code for our experiments is available at [https://github.com/WesternDundrey/Model-X-for-SAEs]. The implementation integrates with the SAELens library and provides:
\begin{itemize}
    \item Complete pipeline from SAE activation collection to knockoff selection
    \item Configurable hyperparameters via YAML or command-line interface
    \item Deterministic execution with fixed random seeds
    \item Detailed artifact logging including knockoff statistics and selected features
\end{itemize}

We use publicly available models (Pythia-70M) and datasets (GLUE SST-2), ensuring full reproducibility of our results.

\bibliographystyle{plainnat}

\end{document}